\documentclass[a4paper, 10pt, conference]{ieeeconf}      
\usepackage{FG2020}

\FGfinalcopy 

\IEEEoverridecommandlockouts                              
\usepackage{balance}
\usepackage{graphics} 
\usepackage{epsfig} 
\usepackage{mathptmx} 
\usepackage{times} 
\usepackage{amsmath} 
\usepackage{amssymb}  
\usepackage[bookmarks=false]{hyperref}
\usepackage{booktabs,tabularx}
\newcolumntype{C}{>{\centering\arraybackslash}X} 
\usepackage[vlined,ruled,linesnumbered]{algorithm2e}

\newlength\mylen

\newcommand{\Tr}{\mathsf{\scriptstyle T}}

\renewcommand{\epsilon}{\varepsilon}

\def\FGPaperID{4} 

\title{\LARGE \bf
Achieving Better Kinship Recognition Through Better Baseline
}

\author{\parbox{16cm}{\centering
    {\large Andrei Shadrikov}\\
    {\normalsize
    Independent Researcher\\}}
    \thanks{This work was not supported by any organization.}
}

\begin{document}

\ifFGfinal
\thispagestyle{empty}
\pagestyle{empty}
\else
\author{Anonymous FG2020 submission\\ Paper ID \FGPaperID \\}
\pagestyle{plain}
\fi
\maketitle

\begin{abstract}

Recognizing blood relations using face images can be seen as an application of face recognition systems with additional restrictions.
These restrictions proved to be difficult to deal with, however, recent advancements in face verification show that there is still much to gain using more data and novel ideas.
As a result face recognition is a great source domain from which we can transfer the knowledge to get better performance in kinship recognition as a source domain.
We present new baseline for automatic kinship recognition task and relatives search based on RetinaFace\cite{deng2019retinaface} for face registration and ArcFace\cite{deng2018arcface} face verification model. 
With the approach described above as the foundation, we constructed a pipeline that achieved state-of-the-art performance on two tracks in the recent Recognizing Families In the Wild Data Challenge.


\end{abstract}

\section{INTRODUCTION} \label{intro}

Face recognition using neural networks and RGB image data is still on the rise thanks to the abundance of raw data on the Internet, more public data~\cite{guo2016ms,maze2018iarpa,cao2018vggface2}, and popular computational frameworks being open source.
Algorithms in this field reach new records~\cite{Grother2018,Grother2019} both in quality and resource consumption which were unimaginable several years before.

Automatic kinship recognition using visual information is very similar to face verification and thanks to growing Families In the Wild (FIW)~\cite{robinson2016fiw} dataset is getting more attention in the research community~\cite{robinson2018visual}.

Annual Recognizing Families In the Wild (RFIW) Data Challenge tries to bring the difficult problem closer to have a solution applicable to the real-world tasks. And every year contestants show new approaches based on metric learning~\cite{li2017kinnet,nandy2019kinship}, ensembling different facial features~\cite{laiadi2019kinship} and the introduction of additional data pre-processing~\cite{Aspandi2019}.
We decided to take a step back and start with a better baseline model and then gradually apply different approaches to improve performance.
To our surprise, basic fine-tuning happened to be enough to achieve better performance than other competitors.
In this work we make three main contributions:
\begin{enumerate}
    \item We propose a new baseline for evaluating kinship recognition methods, which is based on recent advancements in the face recognition field.
    \item We designed a pipeline for the fine-tuning face verification models for visual kinship recognition task based on the new baseline.
    \item We show that our approach\footnote{We encourage you to download our implementation and test it: \url{https://github.com/vuvko/fitw2020}} achieves the best performance in  Recognizing  Families  In the  Wild  Data Challenge (Tracks 1\&3).
\end{enumerate}


The rest of this paper is organized as follows: in section~\ref{related} we briefly review existing methods that are useful for our task of kinship recognition and family search, in section~\ref{pipeline} we explain our proposed baseline and pipeline, in section~\ref{experiments} we evaluate our approach on RFIW challenge dataset, in section~\ref{conclusion} we conclude our work and discuss further possible improvements.

\section{RELATED WORK} \label{related}


Our task is to recognize a binary feature (kin, non-kin) given two face images.
There are several different approaches we can use to tackle this problem: from hand-crafted features to generative methods for data augmentation.
But we will focus on closer work in similar fields.









Convolutional neural networks (CNN) became the backbone in the variety of methods in computer vision tasks.
Since the introduction of AlexNet~\cite{krizhevsky2012imagenet} hand-crafted features had been quickly replaced by better and sometimes even faster algorithms based on CNNs.

\textbf{Face recognition} is an example of a fast-moving field in computer vision.
The progress is due to novel datasets~\cite{guo2016ms,cao2018vggface2,maze2018iarpa} for training and evaluation, introduction of different architectures and learning methods~\cite{schroff2015facenet,Liu_2017_CVPR,wang2018cosface,deng2018arcface}, more data challenges~\cite{kemelmacher2016megaface,deng2019lightweight} and industrial interest~\cite{Grother2019}.
Similarity with visual kinship recognition tasks makes the face recognition models a good candidate for fine-tuning for our task.
And it was shown~\cite{robinson2018visual} that better base models can achieve much higher performance than carefully tuned older ones.

\textbf{Image retrieval} also had a success incorporating CNNs.
The deep representation from the penultimate layer was shown~\cite{donahue2014decaf,babenko2015aggregating} to be a good feature extractor.
Typically methods were tuned for retrieval task with metric learning~\cite{zheng2017sift} but classification approach using proxies~\cite{movshovitz2017no,zhai2018classification} became a new promising design.

\section{PROPOSED PIPELINE} \label{pipeline}

In~\cite{robinson2018visual} fine-tuned SphereFace~\cite{Liu_2017_CVPR} was proposed as the best baseline benchmark for our task.
However, better algorithms were proposed in recent years.
ArcFace~\cite{deng2018arcface} achieved better performance in face verification and was widely recognized~\footnote{Original implementation: \url{https://github.com/deepinsight/insightface}} in Github community with several re-implementations easily obtainable in every popular framework.
Thus, we tried to use it as a new baseline and foundation for our design.

\subsection{Extracting Face Embeddings} \label{detect}

We could use images from FIW dataset without the special preparations to obtain the image embedding, but facial recognition models work differently based on the different face alignment techniques that were used during their training.
Moreover, better face detection and registration further improve model performance~\cite{deng2019retinaface}.
Knowing this, we re-detected faces in the challenge's dataset and aligned them with landmarks from the RetinaFace~\cite{deng2019retinaface} detector.
At this step, some faces were not detected with the selected confidence threshold and were removed from the training and validation set.
For the test set, such images were just resized to fit into the face recognition model.
There were a total of $4$ images removed from the training set, $3$ images from validation, and $4$ problematic images that occurred in the test set.

The ArcFace~\cite{deng2018arcface} model was used for features extraction.
This model was pre-trained on cleaned MS-Celeb-1M~\cite{guo2016ms} dataset~\footnote{Clean dataset can be downloaded from \url{https://github.com/deepinsight/insightface/wiki/Dataset-Zoo}} and has the embedding dimension of 512.

To compare two images $u_i$ and $u_j$ we used cosine distance between their computed embeddings $x_i$ and $x_j$: 

\begin{equation}
\label{eq:cosine}
    d(u_i,\ u_j) =  \frac{ x_i^\Tr x_j }{ \| x_i \|_2 \| x_j \|_2 }
\end{equation}

\subsection{Transfer Learning} \label{classification}

The main difference between face and kinship recognition tasks is in the relative distance between different people.
While face recognition cares mostly about pictures of the same person being close in embedding space, kinship recognition is trying to achieve a much harder task.
In the later different people must be closer to each other than the other groups of people, ideally forming family clusters.
The difference in the available labeled data and similarity between the tasks makes face recognition a great source domain for transferring to the kinship recognition domain.

In the previous iterations of RFIW, the metric learning was used as a transfer learning approach and achieved a great performance~\cite{li2017kinnet}.
However, it requires pairs of images to be carefully sampled to confidently estimate the distribution of all distances in the metric space.
Another concern is that it requires a significant amount of time to train the final model.
Given the information for each person about their family association, we can construct a family classification problem similar to the recent methods in face recognition~\cite{Liu_2017_CVPR,deng2018arcface,wang2018cosface} and metric learning for image retrieval~\cite{movshovitz2017no,zhai2018classification}.
With this our loss function looks like:

\begin{equation}
    L_{class} = - \frac{1}{B} \sum_{i=1}^{B} \log \frac{e^{W_{y_i}^\Tr x_i + b_{y_i}}}{ \sum_{j=1}^{N} e^{ W_{j}^\Tr x_i + b_{j} } } ,
\end{equation}

where $B$ is the batch size, $N$ is the number of families, $x_i$ is the image embedding of a member of the $y_i$-th family, $W$ is the weight matrix (with $W_j$ denoting it's $j$-th column) and $b$ is the bias from classification layer.

\subsection{Forming Validation Pairs} \label{forming_pairs}

In track 1 of RFIW Data Challenge images are divided between families.
As the distribution between persons and between families in challenge's data is non-uniform, we need to be careful with sampling the pairs, as validating model offline is crucial given the limited number of submissions.
We sampled $5\ 000$ positive and $5\ 000$ negative pairs selected uniformly between all families from the validation set using algorithm~\ref{alg:validation_pairs}.

\begin{algorithm}[t]
    \SetAlgoLined
    \small
    \DontPrintSemicolon
    \LinesNumbered
    \KwIn{$N$ number of families\newline
          $k$ number of image pairs to sample}
    \KwResult{$k$ sampled positive and $k$ sampled negative pairs}
    
    $positive\_pairs \leftarrow [\ ]$ \;
    $negative\_pairs \leftarrow [\ ]$ \;
    $A \leftarrow uniform(1,\ N,\ size=k)$\, \# anchor families\;
    \ForAll{$i = 1,\dots,k$}{
        $O \leftarrow $ families without $A_i$ \;
        $anchor\_person \leftarrow $ sample random member from $A_i$ \;
        $positive\_person \leftarrow $ sample other member from $A_i$\;
        $negative\_person \leftarrow $ sample random member from $O$ \;
        $anchor\_face \leftarrow $ sample random image from $anchor\_person$ \;
        $positive\_face \leftarrow $ sample random image from $positive\_person$ \;
        $negative\_face \leftarrow $ sample random image from $negative\_person$ \;
        append $(anchor\_face,positive\_face)$ to $positive\_pairs$\;
        append $(anchor\_face,negative\_face)$ to $negative\_pairs$\;
    }
    
    \caption{Sampling validation image pairs}
    \label{alg:validation_pairs}
\end{algorithm}

Using this approach validates our model without the issue of popular families that have lots of images.
We used AUC ROC metric on the validation pairs (see fig. \ref{fig:verification_val}) for choosing the best model for submission
Another problem occurs though: as we select uniformly between the families, families with low member count are given higher priority.
We chose to resolve this issue with a higher binarization threshold.

\subsection{Choosing Binarization Threshold} \label{binarization}

Given a comparison of two image embeddings using cosine distance~(\ref{eq:cosine}) we need some binarization function to get the needed result.
In our case, a simple binarization by threshold was used.
The threshold can be chosen based on the trade-off between the false positive rate and the true positive rate.
As we had no prior knowledge of how the test pairs were selected we chose target false positive rate based on our three submissions for the final phase.

Given the information about the pair's possible kind of kinship relation, we could have chosen the threshold for every kind separately.
Alas, we could only submit $15$ results and some kinds of relations had a low number of pairs to confidently select threshold.
We further discuss this in section~\ref{verification_results}.

\subsection{Using for Retrieval} \label{retrieval}

The retrieval task can be reduced to a series of verification tasks.
Every probe image is matched with every gallery image to construct a retrieval matrix (every row contains a retrieval result for the probe image).
Our task also has more than one image for every probe person, and we can solve this in different ways.
One way is creating an aggregated feature for each probe person.
We can do this by averaging all the embeddings from their images.
Thus we reduced our task to a single feature per probe and can form a retrieval matrix.

The other way is using all matching results from all images.
We need to add aggregation function $g(\cdot)$ that would consolidate distances between images to adapt to this.
Given a gallery image $u$ and a probe person $P$ with $m$ images $\langle p_1,\ \dots,\ p_m \rangle$ we can sort the gallery images using the distance:

\begin{equation}
    d_{aggr}(P,\ u) = g\left(\ \langle d(p_1, u),\ \dots,\ d(p_m, u) \rangle\ \right)
\end{equation}

Aggregation function should take a vector with an arbitrary number of elements and return a single real number.
There are different options for such function but we tested only \textit{mean} and \textit{max} in this challenge.

With this approach, we need to compare embeddings for every probe image with every gallery image.
It can be difficult to compute with reasonable resources when there is a large number of images per person.
There is a variety of approaches~\cite{zheng2017sift} for that purpose: from reducing latent space~\cite{jegou2010product,baranchuk2018revisiting} to constructing special structures~\cite{malkov2018efficient} but in this work we will only show (see \ref{retrieval_results}) that getting a mean embedding for a person is comparable with searching using the aggregated function $g(\cdot)$.

\section{EXPERIMENTS} \label{experiments}

\begin{table*}
    \caption{Comparison of methods on test set of RFIW2020}   
    \label{tab:verification_test}
    \begin{tabularx}{\textwidth}{@{}l*{11}{C}c@{}}
         \toprule
         method                        & MD & MS &SIBS& SS & BB & FD & FS &GFGD&GFGS&GMGD&GMGS& Average \\
         \midrule
         pretrained                    &0.73&0.69&0.70&0.72&0.69&0.67&0.71&0.62&0.56&0.65&0.51&  0.70 \\
         \ +classification             &0.78&0.73&0.77&0.79&0.78&0.73&0.79&0.75&0.69&0.72&0.57&  0.77 \\
         \ +normalization              &\textbf{0.78}&0.74&\textbf{0.77}&\textbf{0.80}&\textbf{0.80}&0.75&0.81&0.78&0.69&\textbf{0.76}&0.58&  \textbf{0.78} \\
         \ +different thresholds       &\textbf{0.78}&0.74&\textbf{0.77}&\textbf{0.80}&\textbf{0.80}&0.75&0.81&0.78&0.69&\textbf{0.76}&0.60&  \textbf{0.78} \\
         \addlinespace
         \textbf{2nd} place                  &0.75&0.74&0.75&0.77&0.77&0.74&0.81&0.72&\textbf{0.73}&0.67&\textbf{0.68}&  0.76 \\
         \textbf{3rd} place                   &0.75&\textbf{0.75}&0.72&0.74&0.75&\textbf{0.76}&\textbf{0.82}&\textbf{0.79}&0.69&\textbf{0.76}&0.67&  0.76 \\
         \bottomrule
    \end{tabularx}
\end{table*}


\begin{figure}[t]
    \centering
    \includegraphics[width = 0.4 \textwidth]{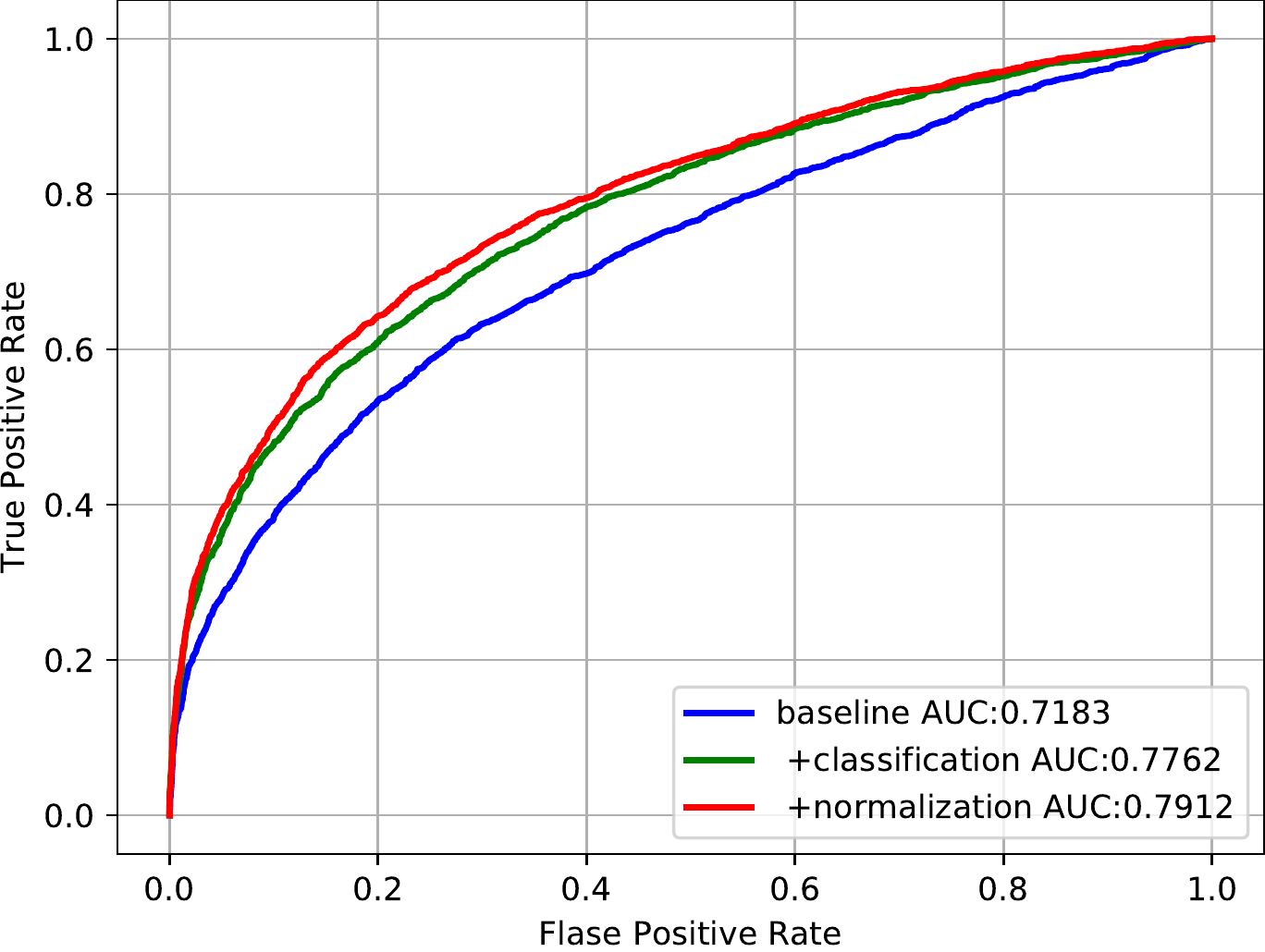}
    \caption{Receiver operating characteristic curve for approaches tested on our validation. We saw the correlation between area under the curve (AUC) score and leaderbord placement, and because of that algorithms were first evaluated offline before submitting. Only better variations were submitted.}
    \label{fig:verification_val}
\end{figure}

\subsection{Recognizing Families In the Wild Data Challenge}

The Recognizing Families In the Wild Data Challenge (RFIW2020) focuses on determining blood relations based on visual facial similarities.
For that, it has $20867$ images from $763$ families for training and validation sets.
This year there were a total of three tracks: one-to-one kinship verification (track 1), two-to-one kinship verification (track 2), and family search and retrieval (track 3).
Our team chose to participate in tracks~1\&3 so we will focus only on them.

In track 1 there were $39743$ pairs for the final testing.
Methods were evaluated based on the average accuracy of binary classification (kin, non-kin) over all of the testing pairs.
Additional measurements were provided separately for every kinship type.

Track 3 is the image retrieval problem with one family member as a query (or probe) and other family members with distractors as a gallery. There were $190$ probe subjects (each with a different number of images) and $3897$ images in the gallery.
Methods were evaluated based on mean average precision (mAP) and rank@k metrics.

\subsection{Implementation Details}

We used Mxnet~\cite{chen2015mxnet} for the implementation of our pipeline.
For detection and feature extraction \textit{insightface} python package was used.
In particular, \textit{retinaface\_r50\_v1} which is RetinaFace implementation with ResNet50 as the backbone and \textit{arcface\_r100\_v1} which is modified ResNet101 trained with ArcFace loss on cleaned  MS-Celeb-1M  dataset.

Re-detected and aligned (as described in \ref{detect}) faces were given to the feature extractor model to obtain image embeddings.
Performance of this approach (\textit{pretrained} on fig. \ref{fig:verification_val}) was used as a baseline to test our hypotheses.

First, we tried to add a simple classification layer and finetune the whole model on the train set with stochastic gradient descent with base learning rate of $0.0001$, momentum $0.9$, linear warmup for first $200$ batches of size $64$, linear cooldown for last $400$ batches, multiplying learning rate by $0.75$ on epochs $8,\ 14,\ 25,\ 35,\ 40$ and gradient clipping $1.5$ for $50$ epochs.
Random color jitter and random lightning with parameter $0.15$ were used for the data augmentation.
No random cropping or similar technique was used to not confuse the model that was trained on similar aligned images.
After that, we added $L_2$ normalization of the embeddings and retrained the model starting with pre-trained weights.
Performance of these two models on our sampled validation pairs (see~\ref{forming_pairs}) can be seen in fig.~\ref{fig:verification_val}.

\subsection{Verification Results} \label{verification_results}

We needed to binarize our predictions to submit for test verification on track 1.
We chose threshold such that we had true positive rate (TPR) of $0.75$ on our sampled validation images for our first submission (\textit{pretrained} in table~\ref{tab:verification_test}).
Next submissions were tested with different thresholds and between several strategies, the one that showed the best average performance was to choose the threshold such that the method would have a false positive rate (FPR) of $0.2$.
Other entries in table~\ref{tab:verification_test} are provided with that strategy of choosing the binarization threshold.

We tested the simple fine-tuning using a classification layer (\textit{+classification}) with a similar approach where the embeddings are normalized to have a unit $L_2$ norm before the classification layer (\textit{+normalization}).
Both on our validation data and test set the second approach was superior.
This indicates that consistency with our cosine distance metric that we use for image comparison is crucial for fine-tuning the model for kinship verification.

From comparison table~\ref{tab:verification_test} we can see that our approach performs poorly on grandparents-grandchildren type of kinship because there is a small number of images with this type of relationship in the training set, but we can mitigate this bias through a different threshold for every kind of relationship.
Sadly, we could not test this idea due to the lack of time, but we provide a proof of this hypothesis with \textit{+different thresholds} submission where we improved the performance for grandmother-grandson relationship by lowering binarization threshold.

We should note that though our approach scores first on average it is not by a far margin and mostly due to our great performance on sibling pairs.
Having an $0.78$ average accuracy at best, automatic kinship recognition still needs to be improved to be considered for usage in real-world applications.
For the reference, the best face verification models perform with FPR around $10^{-6}$~\cite{Grother2019}.

\subsection{Retrieval Results} \label{retrieval_results}

\begin{table}[t]
    \caption{Comparison of retrieval methods on test set of RFIW2020}   
    \label{tab:retrieval_test}
    \begin{tabularx}{\columnwidth}{l*{3}{C}}
         \toprule
         method                        & mAP & Rank@K & Average \\
         \midrule
         pretrained                    &0.163&0.53    & 0.34 \\
         \ +norm+class                 &0.192&0.56    & 0.38 \\
         \ +mean aggregation           &\textbf{0.193}&0.57    & 0.38 \\
         \ +max aggregation            &0.179&\textbf{0.60}    & \textbf{0.39} \\
         \addlinespace
         \textbf{2nd} place            &0.08 &0.38    & 0.23  \\
         \textbf{3rd} place            &0.08 &0.36    & 0.23  \\
         \textbf{4th} place            &0.06 &0.32    & 0.19  \\
         \bottomrule
    \end{tabularx}
\end{table}

In track 3 we needed to aggregate several embeddings per probe person to rank gallery images.
We used average consolidated embedding for our baseline submission \textit{pretrained} and compared it to every gallery image using the cosine metric to get a resulting retrieval matrix.
The same procedure was used with our best model from track 1 (\textit{+norm+class}) and we can see that our approach gives improvements not only to the verification task but also to the retrieval.
Then we compared this search method with different aggregation functions $g(\cdot)$ (see~\ref{retrieval}).
We can see that averaging embeddings for probe subjects perform worse than searching using all available embeddings with aggregation function but is still comparable.
Furthermore, we can see that \textit{max} aggregation function, which searches for the image from the gallery that is closest to any of the query images, has a higher rank@K metric than \textit{mean} aggregation but lower mAP.

Table~\ref{tab:retrieval_test} shows that even the pre-trained ArcFace model performs much better than the other competitors and our pipeline improved this performance even further.
But even such great performance is too low for trying to use this pipeline in a real-world scenario.



\section{CONCLUSIONS AND FUTURE WORKS} \label{conclusion}

In this work we show that using better face verification models is crucial for improving kinship recognition due to more available data.
We presented the new baseline for kinship verification and retrieval tasks, which is based on more accurate face recognition model than the previous baseline.
Furthermore our designed pipeline for the verification task improved this result and achieved the best performance in the recent challenge.


In future work we plan to provide a more thorough analysis of methods suitable for the automatic kinship recognition task.
Different feature extractors and ensembling are the most promising next steps from our perspective.

\section{ACKNOWLEDGMENTS}

The author would like to thank Nikolai Amiantov, Konstantin Aleshkin, and Anastasia Belikova for the helpful discussion in preparing this publication, all reviewers for their valuable comments, and the competition organizers for the opportunity to show this work.



\balance
\bibliographystyle{unsrt}
\bibliography{literature}

\end{document}